\documentclass[letterpaper, 12pt]{article}
\usepackage[utf8]{inputenc} 
\usepackage{graphicx} 
\usepackage{ifpdf}
\usepackage{xcolor}
\usepackage{ amssymb }
\DeclareGraphicsExtensions{.pdf}
\usepackage{fullpage}
\usepackage[all]{xy}
\usepackage[english]{babel}

\usepackage[cmex10]{amsmath}
\usepackage{subcaption}
\usepackage[most]{tcolorbox}
\usepackage{wrapfig}
\usepackage{lipsum}
\usepackage{float}
\usepackage{siunitx}
\usepackage{url}
\usepackage[backend=biber,style=apa]{biblatex}
\DeclareLanguageMapping{english}{english-apa}

\usepackage{csquotes}
\graphicspath{Enunciados/}
\usepackage{csquotes}
\usepackage{siunitx} 
\usepackage{multicol}
\usepackage{listings}
\usepackage{hyperref}
\usepackage{listings}

\definecolor{codegray}{gray}{0.95}

\lstdefinestyle{mystyle}{
    backgroundcolor=\color{codegray},
    basicstyle=\ttfamily\footnotesize,
    breaklines=true,
    frame=single,
    captionpos=b,
    numbers=left,
    numberstyle=\tiny\color{gray},
    keywordstyle=\color{blue},
    commentstyle=\color{gray},
    stringstyle=\color{red},
    showstringspaces=false
}
\lstdefinestyle{cppStyle}{
    language=C++,      
    basicstyle=\ttfamily\color[HTML]{A7ADAE}, 
    keywordstyle=\color[HTML]{A67ABB}, 
    commentstyle=\color[HTML]{4C8B52}, 
    stringstyle=\color[HTML]{C67C47}, 
    numbers=left,      
    numberstyle=\tiny\color[HTML]{A7ADAE}, 
    stepnumber=1,      
    numbersep=5pt,     
    backgroundcolor=\color[HTML]{1F1F1F}, 
    showspaces=false,    
    showstringspaces=false, 
    showtabs=false,      
    tabsize=2,         
    breaklines=true,     
}

\lstdefinestyle{htmlStyle}{
    language=HTML,      
    basicstyle=\ttfamily\color[HTML]{A7ADAE}, 
    keywordstyle=\color[HTML]{A67ABB}, 
    commentstyle=\color[HTML]{4C8B52}, 
    stringstyle=\color[HTML]{C67C47}, 
    numbers=left,      
    numberstyle=\tiny\color[HTML]{A7ADAE}, 
    stepnumber=1,      
    numbersep=5pt,     
    backgroundcolor=\color[HTML]{1F1F1F}, 
    showspaces=false,    
    showstringspaces=false, 
    showtabs=false,      
    tabsize=2,         
    breaklines=true,     
}

\usepackage{enumerate} 
\begin{document}
\vspace*{-1cm}
\includegraphics[width=2.2cm]{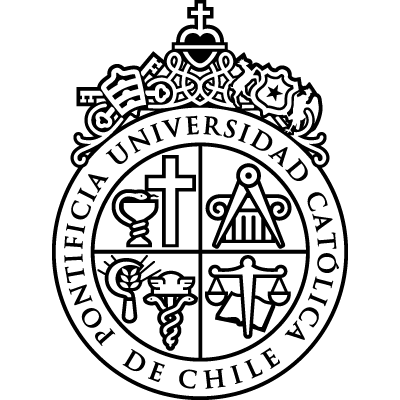}
\vspace*{-2cm}

\hspace*{1.7cm}
  \begin{tabular}{l}
  {\ Pontificia Universidad Católica de Chile.}\\
  {\ School of Engineering}\\
  {\ Directorate for Research and Innovation}\\
  {\ iPre Undergraduate Research Program}\
  \end{tabular}
  \hfill 
\vspace*{4mm}
\begin{center}
{\LARGE\bf Modeling and Control of Magnetic Forces between Microrobots}\\
\vspace*{2mm}

Amelia Fernández Seguel \textsuperscript{a}, Alejandro I. Maass\textsuperscript{b} \\
\end{center}
{\footnotesize
\textsuperscript{a} \textit{Robotic Engineering, School of Engineering, Pontificia Universidad Católica de Chile. 2nd Year, afernandezs3@estudiante.uc.cl} \\
\textsuperscript{b} \textit{Department of Electrical Engineering, School of Engineering, Pontificia Universidad Católica de Chile., Assistant Professor, alejandro.maass@uc.cl}}\\

\hrule

\setlength{\parindent}{0pt}

\vspace*{5pt}
\section*{Abstract}
The independent control of multiple magnetic microrobots under a shared global signal presents critical challenges in biomedical applications such as targeted drug delivery and microsurgeries. Most existing systems only allow all agents to move synchronously, limiting their use in applications that require differentiated actuation.\\

This research aims to design a controller capable of regulating the radial distance between micro-agents using only the angle $\psi$ of a global magnetic field as the actuation parameter, demonstrating great potential for practical applications.\\

The proposed cascade control approach enables faster and more precise adjustment of the inter-agent distance than a proportional controller, while maintaining smooth transitions and avoiding abrupt changes in the orientation of the \textbf{MAGNETIC FIELD}, making it suitable for real-world implementation.\\

To this end, a bibliographic review was first conducted to develop the physical model, considering magnetic dipole–dipole interactions and velocities in \textbf{VISCOUS MEDIA}. Subsequently, a PID controller was designed and implemented to regulate the radial distance, followed by a PD controller in cascade to smooth changes in the field’s orientation.\\

These controllers were simulated in MATLAB, showing that the PID controller reduced the convergence time to the desired radius by approximately 40\%. When adding the second controller, the combined PID+PD scheme achieved smooth angular trajectories within similar timeframes, with fluctuations of only $\pm 5^\circ$.\\

These results validate the feasibility of controlling the radial distance of two microrobots using a shared magnetic field in a fast and precise manner, without abrupt variations in the control angle - an improvement over previously proposed controllers. However, the current model is limited to a 2D environment and two agents, suggesting future research to extend the controller to 3D systems and multiple agents.\\

\textit{\textbf{Keywords:} microrobots, magnetic control, multi-agent systems, PID control, biomedicine.\\}

\hrule

\section{Introduction}
Robotics has transformed numerous scientific fields, and medicine is no exception. From the introduction of automated devices in surgeries to the development of technologies for rehabilitation, the integration of robotics into medical environments has enhanced diagnosis, treatment, and overall healthcare effectiveness by enabling greater precision, reduced invasiveness, and improved clinical outcomes. (\cite{Agrawal2024RoboticsMedical}) \\

One of the most recent examples is “Pill Bot,” an ingestible capsule robot that explores the gastrointestinal tract in real time, simplifying diagnostic procedures (\cite{pillbot}). However, future medical needs demand even smaller and more autonomous systems, which is why the concept of medical robotics is gaining increasing relevance (\cite{conceptorobotica}).\\

Within this context, medical robotics can be classified into three scales (\cite{clasificacion}):

\begin{itemize}
    \item \textbf{Macrorobotics:} Robotic arms and prosthetic devices
    \item \textbf{Microrobotics:} Structures smaller than 1 mm, with cross-disciplinary applications such as minimally invasive interventions 
    \item \textbf{Nanorobotics:} Systems for drug delivery or direct manipulation of molecular structures
\end{itemize}

Unlike macroscopic robots, microrobots are governed by microscale forces (surface tension, chemical interactions) and lack conventional actuators or sensors.  Their control relies on external stimuli such as light, ultrasound, or magnetic fields (\cite{LamkinKennard2018MolecularCellular}). Among these, magnetic control stands out for its safety, ability to generate precise 3D forces, and compatibility with biological tissues. (\cite{Shao2021MagneticControl}) \\

However, the challenge lies in independently controlling multiple magnetic microrobots under a shared global signal. Most current systems can only move all agents synchronously, which limits their application in contexts requiring individual movement. (\cite{salehizadeh2020})\\

This research note addresses the problem of differentiated control in magnetic multi-agent systems, proposing strategies to modulate inter-agent forces and achieve selective actuation. The ultimate goal is to develop a control system applicable to practical scenarios such as targeted drug delivery or microsurgeries, where precision and autonomy among agents are essential.

\section{Initial Research}

\subsection{System Characteristics}
The microrobots are modeled as disks with a radius of $R = \SI{250}{\micro\meter}$ and a \textbf{MAGNETIC MOMENT} $\mathbf{m}$. They operate in honey to limit their degrees of freedom and minimize capillary effects, simplifying the physical model. 

\subsection{Forces}
\subsubsection{Magnetic field of a dipole}
A dipolar agent \textbf{(MAGNETIC DIPOLE)} generates a magnetic field around itself, which is expressed as follows (\cite{fcampomag}):

\begin{equation}
    \label{campomagentico}
    \mathbf{B}_{dip}(\mathbf{r})=\frac{\mu_0}{4\pi}\frac{1}{r^3}[3(\mathbf{m_1} \cdot \hat{\mathbf{r}})\hat{\mathbf{r}} - \mathbf{m}]
\end{equation}

Where:
\begin{itemize}
    \item  $\mathbf{B}_{dip}(\mathbf{r})$ is the magnetic field generated by the agent.
    \item $\mathbf{r}$ is the vector from the source dipole (e.g., agent 1) to the point where the field is evaluated (e.g., agent 2).
    \item $\mu_0$ is the permeability of free space (${4\pi \cdot 10^{-7}}{H/m}$).
    \item $r$ is the distance between agents.
    \item $\hat{\mathbf{r}}$ is a unit vector indicating the direction of $\mathbf{r}$.
    \item $\mathbf{m}$ is the magnetic moment of the agent.
\end{itemize}

\subsubsection{Permeability of the medium}
The interaction between magnetic dipoles occurs through the magnetic field in the surrounding medium, which depends on the medium’s magnetic permeability. This value is given by (\cite{permeabilidad}):

\begin{equation}
    \mu = \mu_0 \mu_r
\end{equation}

Since the agents are submerged in honey, which, being organic, has a \textbf{MAGNETIC PERMEABILITY} close to 1 ($\mu_r \approx 1$), the total permeability is:

\begin{tcolorbox}[colframe=black, colback=white, sharp corners, boxrule=1pt]
\begin{equation}
\label{permeabilidad}
    \mu = \mu_0\cdot1 = \mu_0
\end{equation}
\end{tcolorbox}

Thus, the expression in (\ref{campomagentico}) remains valid.\\

\subsubsection{Forces between microrobots}

Based on these parameters, the force experienced by agent 2 due to the field $\mathbf{B}_{dip}$ (\ref{campomagentico}) can be calculated as (\cite{fuerzacamp}):

\begin{equation}
    \label{F}
    \mathbf{F}=\nabla (\mathbf{m_2} \cdot \mathbf{B_{dip}})
\end{equation}

To simplify control, and following assumptions made in previous studies (\cite{salehizadeh2020}), all magnetic moments of the microrobots are considered aligned with the applied field, denoted as $\mathbf{b_a}$.\\

Given that the agents are identical (and thus have the same magnetic moment), we define:

\begin{equation}
    \mathbf{m}_1 = \mathbf{m}_2 = {\mathbf{m\hat{b}}}_a
\end{equation}

Where $\mathbf{m}_1$ and $\mathbf{m}_2$ correspond to the moments of agents 1 and 2, and ${\mathbf{m\hat{b}}}_a$ represents the magnitude of the moment in the direction of ${\mathbf{{b}}}_a$.\\

Substituting these expressions into (\ref{campomagentico}) and then into (\ref{F}) yields:

\begin{equation}
    \mathbf{F}=\frac{3\mu_0\mathbf{m^2}}{4\pi r ^4}\left[ (\hat{\mathbf{b}}_a \cdot \hat{\mathbf{r}}) \hat{\mathbf{r}} + (\hat{\mathbf{b}}_a \cdot \hat{\mathbf{r}}) \hat{\mathbf{b}}_a + (\hat{\mathbf{b}}_a \cdot \hat{\mathbf{b}}_a) \hat{\mathbf{r}} - 5(\hat{\mathbf{b}}_a \cdot \hat{\mathbf{r}})^2 \hat{\mathbf{r}} \right]
\end{equation}

\subsubsection{Local coordinates}

Since $\hat{\mathbf{b}}_a$ is a unit vector representing the direction of $\mathbf{b_a}$, it follows that $(\hat{\mathbf{b}}_a \cdot \hat{\mathbf{b}}_a = 1)$, allowing the equation to be simplified as:

\begin{equation}
\mathbf{F} = \frac{3 \mu_0 \mathbf{m^2}}{4 \pi r^4} \left[2(\hat{\mathbf{b}}_a \cdot \hat{\mathbf{r}}) \hat{\mathbf{b}}_a +\hat{\mathbf{r}} -5(\hat{\mathbf{b}}_a \cdot \hat{\mathbf{r}})^2 \hat{\mathbf{r}}\right]
\end{equation}

\begin{figure}[H]
    \centering
    \includegraphics[width=0.65\linewidth]{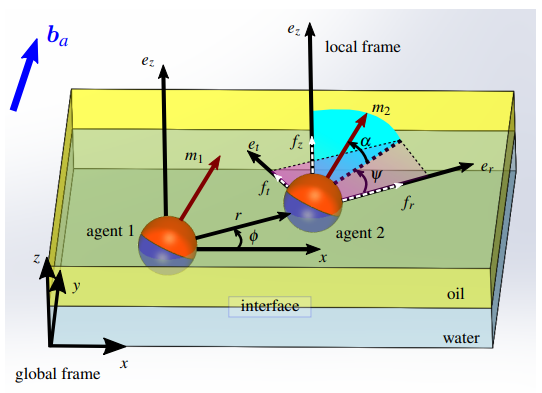}
    \caption{Representation of agent coordinates in a water–oil medium (\cite{salehizadeh2020})}
    \label{fig:fig1}
\end{figure}

To decompose this force into local coordinates, an orthonormal system is defined. The agents are considered to be in a water–oil medium for simplicity (Figure \ref{fig:fig1})

This system is parameterized in spherical coordinates as:

\begin{equation}
\label{campo}
    \hat{\mathbf{b}}_a = \cos \alpha \cos \psi\, \hat{\mathbf{e}}_r + \cos \alpha \sin \psi\, \hat{\mathbf{e}}_t + \sin \alpha\, \hat{\mathbf{e}}_z
\end{equation}

Where:
\begin{itemize}
    \item $\psi$: In-plane angle ($xy$) between the projection of $\mathbf{b_a}$ and the radial vector $\mathbf{\hat{e}_r}$.
    \item $\alpha$: Out-of-plane angle ($z$) between $\mathbf{b_a}$ and its projection on the ($xy$) plane.
    \item $\hat{e}_r$ = $\hat{\mathbf{r}}$: Radial direction.
    \item $\hat{e}_t$: Tangential direction in the plane, orthogonal to $\hat{e}_r$.
    \item $\hat{e}_z$: Direction perpendicular to the plane.
\end{itemize}

A spherical parameterization was used because the agents are spherical.\\

\subsubsection{Minimum magnitude of the applied field}
The magnitude of $\mathbf{b_a}$ must ensure that the dipoles remain aligned with it, without being significantly affected by their mutual proximity. This condition is expressed as:

\begin{equation}
    \label{eq: magnitud minima}
    |\mathbf{b_a}|_{min} = \frac{b-\sqrt{b^2-4ac}}{2a}
\end{equation}

With:
\begin{subequations}
\begin{align}
    a &= \frac{1}{(A_1)^2} - \frac{1}{(A_2)^2} \\
    b &= \frac{\mu_0 \mathbf{m} \left( \frac{4}{(A_1)^2} + \frac{2}{(A_2)^2} - 3 \right)}{2\pi r^3} \\
    c &= \frac{\mu_0^2 \mathbf{m^2} \left( \frac{-4}{(A_1)^2} + \frac{1}{(A_2)^2} + 6 \right)}{16\pi r^6} \\
    A_1 &= \cos\left(\theta_\epsilon + \cos^{-1}(A_2)\right) \\
    A_2 &= \cos(\psi)\cos(\alpha)
\end{align}
\end{subequations}

Here, $\theta_\epsilon$ represents the maximum error angle.  
The minimum magnitude $|\mathbf{b}_a|_{\text{min}}$ ensures that the applied field dominates local interactions, maintaining dipole alignment. The terms $A_1$ and $A_2$ capture the angular dependencies ($\psi$, $\alpha$) and the maximum allowable error $\theta_\epsilon$.

\subsubsection{Simplification under assumptions}

To determine the force components along each direction, $\mathbf{f}_{21}$ is projected onto the corresponding axes.\\

\textbf{Radial component:}

\begin{equation}
    f_r = \mathbf{f}_{21} \cdot \hat{\mathbf{e}}_r = \frac{3 \mu_0 \mathbf{m^2}}{4 \pi r^4} \left[2 (\cos \alpha \cos \psi)^2 + 1 - 5 (\cos \alpha \cos \psi)^2 \right].
\end{equation}

Simplified to: 

\begin{equation}
\label{fuerzaradial}
    f_r = \frac{3 \mu_0 \mathbf{m^2}}{4 \pi r^4} \left[1 - 3 \cos^2 (\alpha) \cos^2( \psi) \right].
\end{equation}

\vspace{5mm}
\textbf{Tangential component:}

\begin{equation}
f_t = \mathbf{f}_{21} \cdot \hat{\mathbf{e}}_t = \frac{3 \mu_0 \mathbf{m^2}}{4 \pi r^4} \left[2 \cos^2 \alpha \cos \psi \sin \psi \right].
\end{equation}

Simplified to: 

\begin{equation}
\label{tangencial}
f_t = \frac{3 \mu_0 \mathbf{m^2}}{4 \pi r^4} \cos^2 \alpha \sin 2\psi.
\end{equation}

\vspace{5mm}
\textbf{Normal component:}

\begin{equation}
f_z = \mathbf{f}_{21} \cdot \hat{\mathbf{e}}_z = \frac{3 \mu_0 \mathbf{m^2}}{4 \pi r^4} \left[2 \cos \alpha \cos \psi \sin \alpha \right].
\end{equation}

Simplified to: 

\begin{equation}
f_z = \frac{3 \mu_0 \mathbf{m^2}}{4 \pi r^4} \sin 2\alpha \cos \psi.
\end{equation}

\vspace{5mm}
Defining $\Omega = \frac{3 \mu_0 \mathbf{m^2}}{4 \pi}$, the resulting expressions can be summarized as:
\begin{tcolorbox}[colframe=black, colback=white, sharp corners, boxrule=1pt]

\begin{subequations}
\renewcommand{\theequation}{13.\arabic{equation}} 
\begin{align}
f_r &= \frac{\Omega}{r^4} \left[1 - 3 \cos^2 \alpha \cos^2 \psi \right],\label{eq:fr} \\
f_t &= \frac{\Omega}{r^4} \cos^2 \alpha \sin 2\psi,\label{eq:ft} \\
f_z &= \frac{\Omega}{r^4} \sin 2\alpha \cos \psi.
\end{align}
\end{subequations}
\vspace{1mm}
\end{tcolorbox}

\vspace{5mm}
The resulting force depends on the relative orientation between the applied magnetic field and the position of the second agent with respect to the first.\\

The equations obtained are consistent with those presented in the reference thesis (\cite{salehizadeh2020}).

\subsection{Velocities}
\subsubsection{Radial velocity}
Given that the microrobots operate in a viscous environment at a microscopic scale, their behavior is governed by \textbf{STOKES' LAW} (\cite{stokes}):

\begin{equation}
    f_m = 6 \pi \mu_m R v
\end{equation}

Where:
\begin{itemize}
    \item $\mu_m$ is the viscosity of the medium,
    \item $R$ is the radius of the agent,
    \item $v$ is the displacement velocity.
\end{itemize}

Solving for velocity:

\begin{equation}
    v = \frac{f_m}{6 \pi \mu_m R}
\end{equation}

And defining $\sigma_{tras} = 6 \pi \mu_m R$ as a constant:

\begin{equation}
    v = \frac{f_m}{\sigma_{tras}}
\end{equation}

This equation can be substituted to obtain the radial velocity:

\begin{equation}
\label{stokesradial}
\dot r = \frac{f_r}{\sigma_{tras}} = \frac{\Omega}{\sigma_{tras} r^4} \left[1 - 3 \cos^2 \alpha \cos^2 \psi \right]
\end{equation}

and defining:

\begin{equation}
    \Omega_t=\frac{\Omega}{\sigma_{tras}}=\frac{\frac{3\mu_0 m^2}{4 \pi}}{6 \pi \mu_m R}=\frac{\mu_0 m^2}{8\pi^2 \mu_m R}
\end{equation}

to obtain
\begin{tcolorbox}[colframe=black, colback=white, sharp corners, boxrule=1pt]
\begin{equation}
\dot r = \frac{\Omega_t}{ r^4} \left[1 - 3 \cos^2 \alpha \cos^2 \psi \right]
\end{equation}
\end{tcolorbox}

\subsubsection{Angular velocity}

In the case of $\dot\phi$, torque must be taken into consideration. The tangential force ($f_t$) generates a torque that makes the agent rotate around the center of mass, this is given by:

\begin{equation}
    \tau_{drag} = f_t \cdot r
\end{equation}

The rotational Stokes' law (\cite{stokesrotacional}) indicates that in viscous fluids (as would be the case with honey), the following holds:

\begin{equation}
\label{angularstokes}
\tau_{drag} = \sigma_{rot} \cdot \dot\phi
\end{equation}

Where $\sigma_{rot} = 8\pi \mu R^3$. By equating these two equations:

\begin{equation}
    \sigma_{rot}\cdot \dot\phi  = f_t \cdot r
\end{equation}

We solve for:
\begin{equation}
    \dot\phi  = \frac{f_t\cdot r}{\sigma_{rot}}
\end{equation}

When replacing with (\ref{eq:ft}):

\begin{equation}
    \dot\phi  = \frac{\Omega \cos^2 (\alpha) \sin (2\psi)}{r^3\cdot\sigma_{rot}}
\end{equation}

\vspace{3mm}

\begin{tcolorbox}[colframe=black, colback=white, sharp corners, boxrule=1pt]

These expressions, by defining $\Omega_t = \frac{\Omega}{\sigma_{tras}}$ and $\Omega_r = \frac{\Omega}{\sigma_{rot}}$, can be summarized as:

\begin{subequations}
\label{valoresconstantes}
\begin{align}
\dot{r} &= \frac{\Omega_t}{r^4} \left[1 - 3 \cos^2 \alpha \cos^2 \psi \right] \label{eq:rdot}, \\
\dot{\phi} &= \frac{\Omega_r}{r^3}[\cos^2 (\alpha) \sin (2\psi)] \label{eq:phidot}.
\end{align}
\end{subequations}

With $\Omega=\frac{3\mu_0 m^2}{4 \pi}$, $\sigma_{\text{tras}}=6\pi \mu_m R$ and $\sigma_{\text{rot}}=8\pi \mu_m R^3$, where $\mu_0=4\pi \cdot 10^{-7} \frac{H}{m}$, $m$ corresponds to the magnetic moment of the agents, $R$ to the radius of the agents, $\mu_m$ to the viscosity of the medium.

\end{tcolorbox}

Table~\ref{tab:parametros} summarizes the key physical constants that will be used for the controller design.

\begin{table}[H]
\centering
\begin{tabular}{|c|c|c|l|}
\hline
\textbf{Constant} & \textbf{Value}         & \textbf{Unit} & \textbf{Description}                 \\
\hline
$\mu_0$         & $4\pi \cdot 10^{-7}$   & H/m         & Permeability of free space         \\
$m$             & $6.545 \cdot 10^{-7}$  & A·m²        & Magnetic moment of the microrobot   \\
$R$             & $250 \cdot 10^{-6}$    & m           & Radius of the microrobot         \\
$\mu_m$         & 0.5                    & Pa·s        & Viscosity of the medium (honey)   \\
$\sigma_{\text{tras}}$ & $6\pi \mu_m R$         & N·s/m       & Translational friction constant \\
$\sigma_{\text{rot}}$  & $8\pi \mu_m R^3$         & N·m·s       & Rotational friction constant   \\
$\psi_s$        & 54.74                  & degrees     & Angle at which RADIAL FORCE $\approx 0$ \\
\hline
\end{tabular}
\caption{Physical parameters of the microrobot and its environment}
\label{tab:parametros}
\end{table}

\section{Control}
The following presents the design, simulation, and validation of the control to regulate the radial distance between agents. First, the problem of \textbf{UNDERACTUATION} (Section 3.1) is analyzed, where a single parameter ($\psi$) controls multiple variables $(r, \phi)$. Then, in Section 3.2, the proportional controller (P) is evaluated, to later develop and implement the improved PID controller (Section 3.3) and its cascade version with PD. Finally, Section 3.4 discusses limitations and future extensions.

\subsection{Underactuated Motion}

Using the information mentioned in Section 2 on velocities, a control can be created to manage the distance at which the agents are located. In this case, only the angle $\psi$ is considered as the control signal, assuming that $\alpha=0^\circ$. 

\begin{subequations}
\label{rpunto}
\begin{align}
\dot{r} &= \frac{\Omega_t}{ r^4} \left[1 -3  \cos^2 (\psi) \right] \label{eq:rdot}, \\ 
\dot{\phi} &= \frac{\Omega_r}{r^3 } [\sin (2\psi)]
\end{align}
\end{subequations}

\begin{figure}[H]
    \centering
    \includegraphics[width=0.5\linewidth]{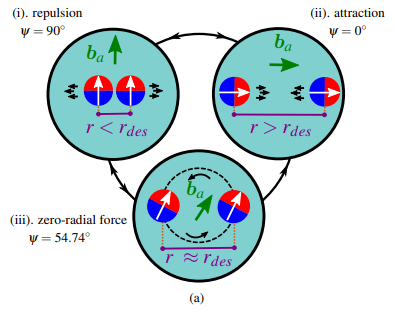}
    \caption{Principle of radial distance control of micro-agents using the angle $\psi$ as the effector parameter (\cite{salehizadeh2020})}
    \label{fig:dipolo}
\end{figure}

Since the agents will be aligned with the field $b_a$, its direction can be used to control the behavior between the agents. Using the polarities, the following three cases are defined, which can be observed in Figure \ref{fig:dipolo}.

\begin{enumerate}
    \item Maximum attraction ($\psi=0^\circ$): When this occurs, the attraction is maximum, with $\dot r=-\frac{2\Omega_t}{r^4}$ and $f_r = -\frac{2\Omega}{r^4}$.
    \item Maximum repulsion ($\psi=90^\circ$): When this occurs, the repulsion is maximum, with $\dot r=\frac{\Omega_t}{r^4}$ and $f_r=\frac{\Omega}{r^4}$.
    \item Zero force ($\psi=54.74^\circ$: At this specific angle, $\dot r= 0.0002 \cdot \frac{\Omega_t}{r^4}$, considered as zero velocity.
\end{enumerate}

Taking the above into account, it is possible to use this information to define a control system. Based on what is stated in (\ref{eq: magnitud minima}), a value $\epsilon$ is defined, representing the minimum allowed distance radius between agents, such that the field $\mathbf{b_a}$ is capable of aligning them with itself and the model is applicable.\\

$r_{min}$ and $r_{max}$ are also defined as limits within which the created controller has greater effectiveness. Outside these values, maximum attraction or repulsion forces are applied.\\

From the aforementioned, a piecewise controller can be generated:

\begin{tcolorbox}[colframe=black, colback=white, sharp corners, boxrule=1pt]

\begin{equation}
\label{controladorporpartes}
    \dot r =
    \begin{cases} 
  \frac{\Omega_t}{r^4} & \epsilon< r<r_{min}   \ \rightarrow \ (\psi = 90°) \\
     f_\text{intermediate}(r, \psi )& r_{min}\leq r\leq r_{max} \ \rightarrow \ (\psi = [0°,90°])\\
     -\frac{2\Omega_t}{r^4} & x>r_{max} \ \rightarrow \ (\psi = 0°)
   \end{cases}
\end{equation}

\end{tcolorbox}

Where $f_\text{intermediate}(r, \psi)$ represents the controller that is yet to be defined, with the objective of adjusting the distance precisely and smoothly within the optimal operating range. \\

This controller will use the angle $\psi$ of the field $\mathbf{b_a}$, considering that the force is null when $\psi = 54.74^\circ$.\\

This specific value is obtained by solving the equation $\dot r = \frac{\Omega_t}{r^4}[1-\cos^2(\psi)] = 0$, implying that $\cos^2(\psi)=\frac{1}{3}$ and therefore $\psi \approx 54.74^\circ $. At this angle, the radial force is nullified, allowing a constant distance between agents to be maintained.\\

The formulas given in (\ref{rpunto}) describe not only the distance $r$, but also the rate of change of $\phi$. Since this is an underactuated system controlled only by $\psi$, the states $r$ and $\phi$ are coupled; when adjusting $\psi$ to correct the radius, $\phi$ will be affected. \\

One possibility to regulate $\phi$ without affecting $r$ is to take advantage of the trigonometric dependence of the forces:

\begin{itemize}
    \item $r$ depends on $\cos ^2(\psi)$, symmetric in $\psi$
    \item $\phi$ depends on $\sin(2\psi)$, asymmetric in $\psi$
\end{itemize}

With this, a Bang-Bang type control can be implemented, that is, a controller that rapidly alternates between two states, to change the sign of $\psi$ (positive and negative) and thus obtain the desired value of $\phi$. To do this, once the desired radius is obtained, the field $b_a$ is applied with $\psi=\pm 54.74^\circ$ to only affect the angle $\phi$ of the agents.

\subsection{Existing P Control}

\subsubsection{Generalities}

In a doctoral thesis by Mohhamad Salehizadeh (\cite{salehizadeh2020}), a proportional controller is applied to regulate the angle of the field $b_a$ in the interval $[r_{min} ,r_{max}]$. For this, a radial error was defined as $\epsilon_r=r-r_{desired}$, allowing a P-type control to be directly applied:

\begin{equation}
    \psi=\psi_s-K||\epsilon_r||
\end{equation}

Where $\psi_s$ corresponds to the angle of $54.74°$, at which the applied force is null. The value of the constant K was apparently configured manually.

It was considered that the given value of $\psi$ be in the interval $[0,90]$ when defining $r_{min}$ and $r_{max} $ indicated in the piecewise controller (\ref{controladorporpartes}).\\

Once the corresponding radius is obtained, the value of the angle in the \textbf{APPLIED MAGNETIC FIELD} ($b_a$) is varied to $\psi = \pm 54.74°$ until the desired value of $\phi$ is obtained. \\

To study the behaviour of this type of controller, simulations were carried out in MATLAB R2024b, replicating its operation under specific conditions. 

\subsubsection{Constants}

First, the value of the constants used was defined. For this, the following calculations were made, considering the values of $\Omega, \sigma_{tras}$ and $\sigma_{rot}$ defined in (\ref{valoresconstantes}).\\

For $\Omega_t $:

\begin{equation}
    \Omega_t = \frac{\Omega}{\phi_{tras}}=\frac{\frac{3\mu_0\mathbf{m^2}}{4\pi}}{6\pi \mu_mR} = \frac{\mu_0 {\mathbf{m^2}}}{8\pi^2\mu_mR}=\frac{4\pi\cdot10^7\frac{H}{m}\mathbf{m^2}}{8\pi^2\mu_m R}= \frac{5 \cdot 10^{-8}}{\pi} \cdot \frac{\mathbf{m^2} \frac{H}{m}}{R\cdot \mu_m}
\end{equation}

For $\Omega_r$:
\begin{equation}
    \Omega_t = \frac{\Omega}{\phi_{tras}}=\frac{\frac{3\mu_0\mathbf{m^2}}{4\pi}}{8\pi \mu_mR^3} = \frac{3\mu_0{\mathbf{m^2}}}{32\pi^2\mu_mR^3}=\frac{3\cdot4\pi\cdot10^{-7}\frac{H}{m}\mathbf{m^2}}{32\pi^2\mu_m R^3}= \frac{3 \cdot 10^{-7}}{8\pi} \cdot \frac{\mathbf{m^2} \frac{H}{m}}{R^3\cdot \mu_m}
\end{equation}

The values of $\mathbf{m^2}$ and $R$ (magnetic moment and radius) will depend on the agent to be controlled, and $\mu_m$ will depend on the viscosity of the medium used. In the case that the same agents proposed in the thesis are used, we will have:

\begin{itemize}
    \item $R = 250 \mu m$
    \item $|M| = 10^4 \frac{A}{m}$
    \item Because the magnetic moment of a dipole is given by $\mathbf{m}=M\cdot V$ where V is the volume, the following calculation is performed: $\mathbf{m}=M\cdot \frac{4}{3}\pi R^3 = 10^4 [\frac{A}{m}] \cdot \frac{4}{3}\ \pi \cdot(250\cdot 10^{-6}[m])^3 \approx6.545 \cdot 10^{-7}[A\cdot m^2]$
    \item $\mu_m$ is not fixed, it corresponds to the viscosity of the medium, which is not explicitly stated in the thesis read
\end{itemize}

When substituting into the simplified equation for $\Omega_t$:

\begin{equation}
    \Omega_t=\frac{5 \cdot 10^{-8}}{\pi} \cdot \frac{\mathbf{m^2} \frac{H}{m}}{R\cdot \mu_m}=\frac{5 \cdot 10^{-8}}{\pi} \cdot \frac{(6.545 \cdot 10^{-7}[A\cdot m^2])^2 [\frac{H}{m}]}{250\cdot 10^{-6} [m] \cdot \mu_m} \approx 2,7271 \cdot 10^{-17} \cdot \frac{1}{\mu_m}[HA^2m^2]
\end{equation}

\vspace{5mm}
And for $\Omega_r$:

\begin{equation}
    \Omega_t =  \frac{3 \cdot 10^{-7}}{8\pi} \cdot \frac{\mathbf{m^2} [ \frac{H}{m}]}{R^3\cdot \mu_m} = \frac{3 \cdot 10^{-7}}{8\pi} \cdot \frac{(6.545 \cdot 10^{-7}[A\cdot m^2])^2[ \frac{H}{m}]}{(250 \cdot 10^{-6})^3\cdot \mu_m} \approx 3.2725 \cdot 10^{-10 } \cdot \frac{1}{\mu_m}[HA^2]
\end{equation}

\subsubsection{Parameters}

Based on these calculations, the parameters used in the simulation were defined: the microrobot radius was considered equal to $250 \cdot 10^{-6},\text{m}$, the magnetic moment was $6.545 \cdot 10^{-7},\text{A}\cdot\text{m}^2$, and the viscosity of the medium was set at $\mu_m = 0.5,\text{Pa}\cdot\text{s}$. Furthermore, the value of $\Omega_t$ was determined as $\Omega_t =\frac{2.7271\cdot 10^{-7}[HA^2m^2]}{\mu_m}$.\\
    
The controller limits and the desired radius were fixed, while the only variable parameter was the initial radius. The values considered were: $r_{min} = 300\cdot 10^{-6},\text{m}$, $r_{max} = 700\cdot 10^{-6}\text{m}$, $k_p = 0.2$ and $r_{des} = 500\cdot 10^{-6}[m]$.\\

Additionally, a saturator was incorporated to restrict the angle calculated by the controller to a range between 0° and 90°, since values outside that interval contradict the system's principles (Figure \ref{fig:dipolo}).

\subsubsection{Simulation}

\begin{figure}[H]
    \centering

    \begin{subfigure}[b]{0.49\textwidth}
        \centering
        \includegraphics[width=\textwidth]{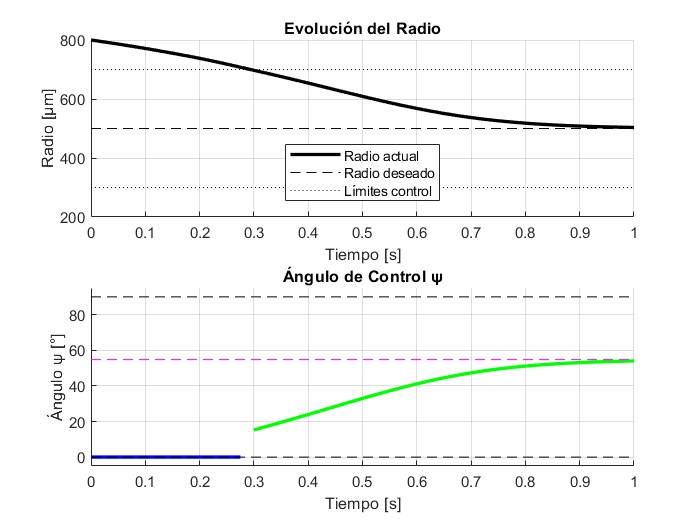}
        \caption{$r_0$ outside the upper range}
        \label{fig:sub1}
    \end{subfigure}
    \hfill
    \begin{subfigure}[b]{0.49\textwidth}
        \centering
        \includegraphics[width=\textwidth]{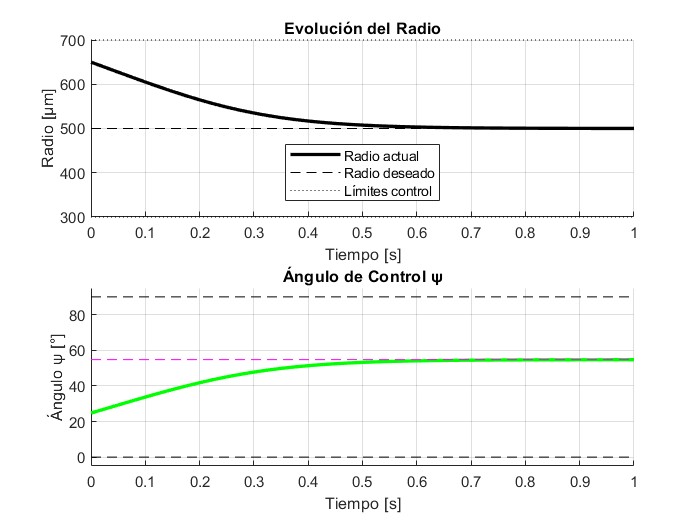}
        \caption{$r_0$ inside the range and greater than $r_{des}$}
        \label{fig:sub2}
    \end{subfigure}

    \vspace{0.5cm} 

    \begin{subfigure}[b]{0.49\textwidth}
        \centering
        \includegraphics[width=\textwidth]{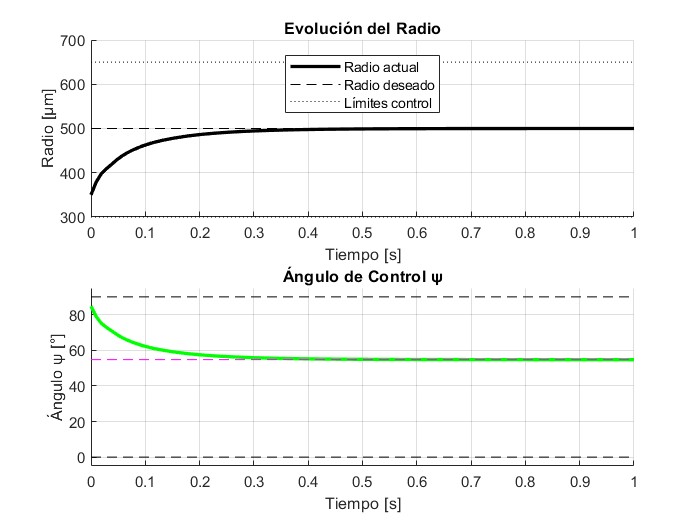}
        \caption{$r_0$ inside the range and less than $r_{des}$}
        \label{fig:sub3}
    \end{subfigure}
    \hfill
    \begin{subfigure}[b]{0.49\textwidth}
        \centering
        \includegraphics[width=\textwidth]{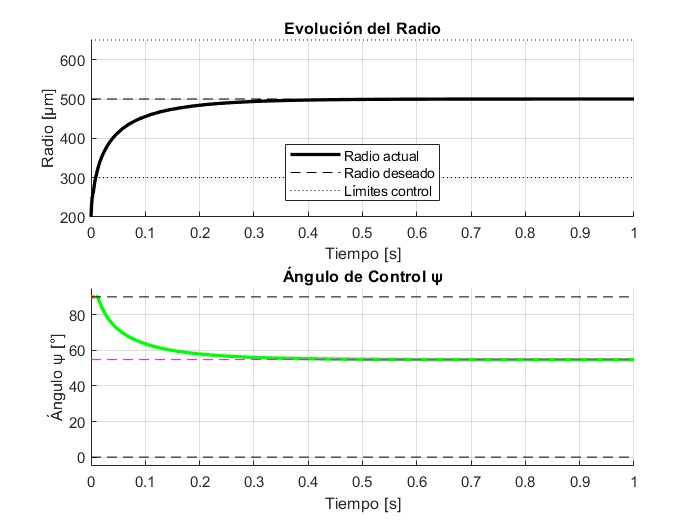}
        \caption{$r_0$ outside the lower range}
        \label{fig:sub4}
    \end{subfigure}

    \caption{Simulation of radial distance change with P-type controller}
    \label{fig:4subfigs}
\end{figure}

Four different cases were analysed according to the value of the initial radius $r_0$:

\begin{itemize}
    \item Outside the upper range, $r_0 = 800\cdot 10^{-6}[m]$ (\ref{fig:sub1})
    \item Inside the range but greater than the desired radius, $r_0 = 650\cdot 10^{-6}[m]$ (\ref{fig:sub2})
    \item Inside the range but less than the desired radius, $r_0 = 350\cdot 10^{-6}[m]$ (\ref{fig:sub3})
    \item Outside the lower range, $r_0 = 200\cdot 10^{-6}[m]$ (\ref{fig:sub4})
\end{itemize}

With Figure \ref{fig:4subfigs}, it can be observed that, although the system does not present steady-state error and the radius and angle curves are smooth, without excessive oscillations, the time it takes to reach the desired value is considerably high in proportion to the distance variation.

\subsection{Proposed Controller}
\subsubsection{PID}

To achieve a smoother and more efficient control response, a \textbf{PID controller} was initially proposed. The goal was to reach the desired radius more rapidly while maintaining the precision already achieved with the proportional controller.\\

For this purpose, the code previously used in the proportional controller simulation was adapted to include a linearized PID controller (Appendix A.\ref{lst:codigoPID}), with the following terms:

\begin{equation}
\psi_p = k_p \cdot error_r \\
\psi_i = \psi_i + k_i \cdot error_r \\
\psi_d = k_d \cdot (error_r - error_{previous})
\end{equation}

Considering that the base angle—where the applied force is practically null according to (\ref{eq:rdot}) - corresponds to $54.74^{\circ}$, the total control angle was defined as:

\begin{equation}
\psi = 54.74 - (\psi_p + \psi_i + \psi_d)
\end{equation}

The parameters $k_p$, $k_i$, and $k_d$ were manually tuned. The process began by maintaining the same $k_p$ as in the proportional controller simulation, then progressively increasing the integral and derivative terms until a stable and smooth response was obtained. \\

\begin{figure}[H]
    \centering
    \begin{subfigure}[b]{0.49\textwidth}
        \centering
        \includegraphics[width=\textwidth]{imagenes/rango_fuera_mayor_P.jpg}
        \caption{Simulation with proportional control}
        \label{fig:5a}
    \end{subfigure}
    \hfill
    \begin{subfigure}[b]{0.49\textwidth}
        \centering
        \includegraphics[width=\textwidth]{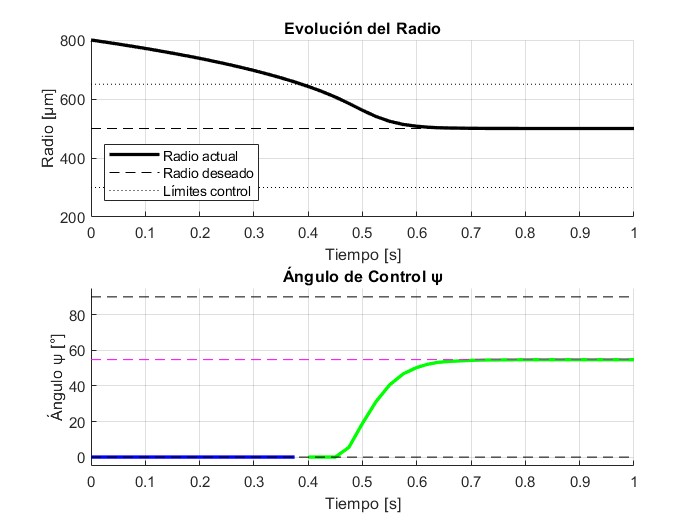}
        \caption{Simulation with PID control}
        \label{fig:5b}
    \end{subfigure}
    \caption{Comparison of controllers with $r_0$ outside the upper range}
    \label{fig:fueradelrangosuperior}
\end{figure}

\begin{figure}[H]
    \centering
    \begin{subfigure}[b]{0.49\textwidth}
        \centering
        \includegraphics[width=\textwidth]{imagenes/dentro_rango_mayor_P.jpg}
        \caption{Simulation with proportional control}
    \end{subfigure}
    \hfill
    \begin{subfigure}[b]{0.49\textwidth}
        \centering
        \includegraphics[width=\textwidth]{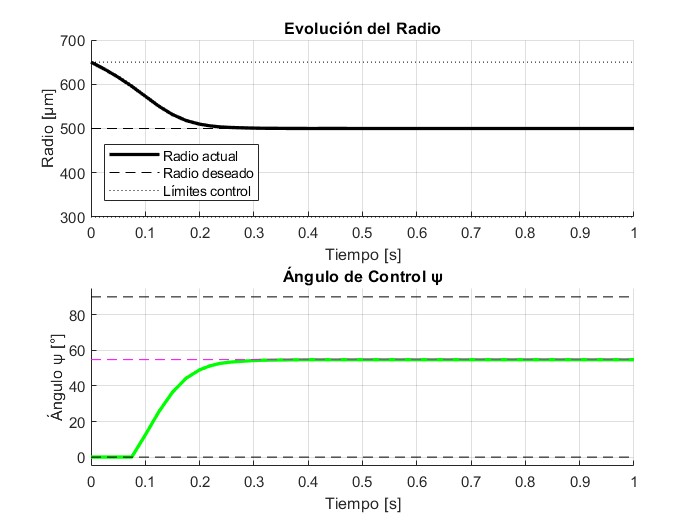}
        \caption{Simulation with PID control}
    \end{subfigure}
    \caption{Comparison of controllers with $r_0$ inside the range and greater than desired}
    \label{fig:dentrodelrangosuperior}
\end{figure}

\begin{figure}[H]
    \centering
    \begin{subfigure}[b]{0.49\textwidth}
        \centering
        \includegraphics[width=\textwidth]{imagenes/dentro_rango_menor_P.jpg}
        \caption{Simulation with proportional control}
    \end{subfigure}
    \hfill
    \begin{subfigure}[b]{0.49\textwidth}
        \centering
        \includegraphics[width=\textwidth]{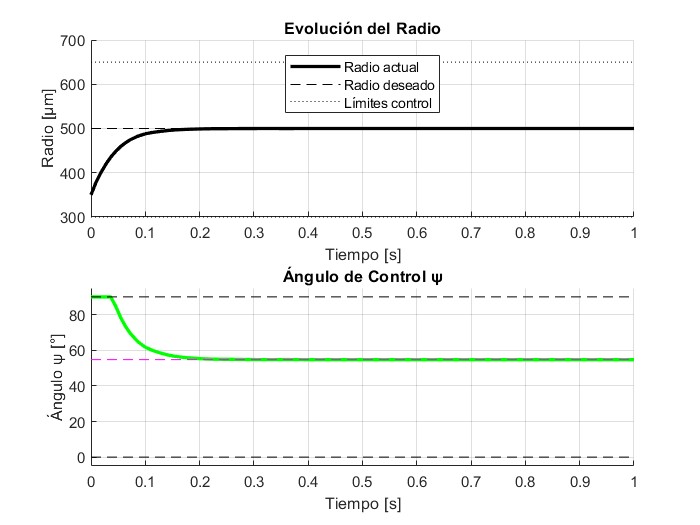}
        \caption{Simulation with PID control}
    \end{subfigure}
    \caption{Comparison of controllers with $r_0$ inside the range and less than desired}
    \label{fig:dentrodel rango}
\end{figure}

\begin{figure}[H]
    \centering
    \begin{subfigure}[b]{0.49\textwidth}
        \centering
        \includegraphics[width=\textwidth]{imagenes/fuera_rango_menor_P.jpg}
        \caption{Simulation with proportional control}
    \end{subfigure}
    \hfill
    \begin{subfigure}[b]{0.49\textwidth}
        \centering
        \includegraphics[width=\textwidth]{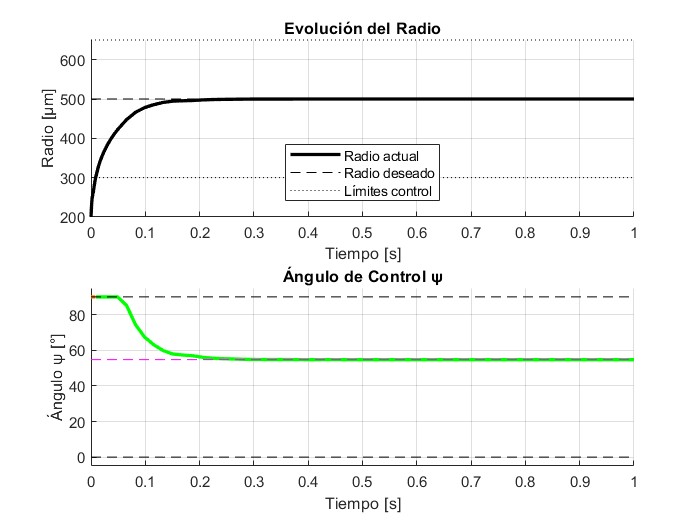}
        \caption{Simulation with PID control}
    \end{subfigure}
    \caption{Comparison of controllers with $r_0$ outside the lower range}
    \label{fig:fueradelranginferior}
\end{figure}

The results were compared under the same conditions as those described in Figure \ref{fig:4subfigs}, but using both the proportional and PID controllers. These comparisons are presented in Figures \ref{fig:fueradelrangosuperior}, \ref{fig:dentrodelrangosuperior}, \ref{fig:dentrodel rango}, and \ref{fig:fueradelranginferior}.

The results show that the PID controller achieves the desired radius faster without generating oscillations that significantly affect either the angle or the radius, demonstrating a clear improvement over proportional control. Furthermore, note that at the beginning of Figures \ref{fig:5a} and \ref{fig:5b}, the blue line in the angle plot indicates that the value computed by the controller temporarily exceeds the limits defined by $r_{min}$ and $r_{max}$ - a situation handled by the previously implemented saturator.\\

The visible jump in this same simulation arises because the plot is in discrete time; those milliseconds represent the moment when the saturator stops acting and only the controller is active.\\

In these tests, a single value for the initial and desired radius was used. To assess system performance under dynamic target variations, an additional simulation was conducted (Appendix A.\ref{lst:codigoPIDmultiple}), where the desired radius changes every second. The initial radius was $r_0 = 800\cdot10^{-6}[m]$, with the following sequence of targets:
\[
50\cdot10^{-6}[m],\ 400\cdot10^{-6}[m],\ 600\cdot10^{-6}[m],\ 550\cdot10^{-6}[m],\ 450\cdot10^{-6}[m]
\]

\begin{figure}[H]
    \centering
    \includegraphics[width=\linewidth]{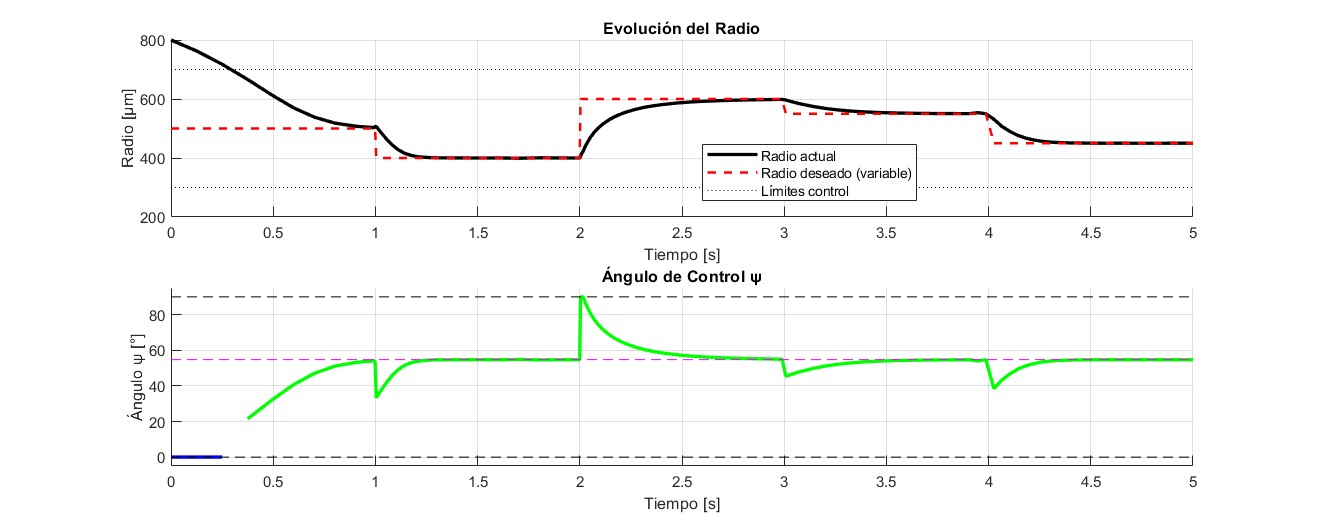}
    \caption{Simulation of proportional control applied to multiple target radii}
    \label{fig:radiosmultiplesP}
\end{figure}

\begin{figure}[H]
    \centering
    \includegraphics[width=\linewidth]{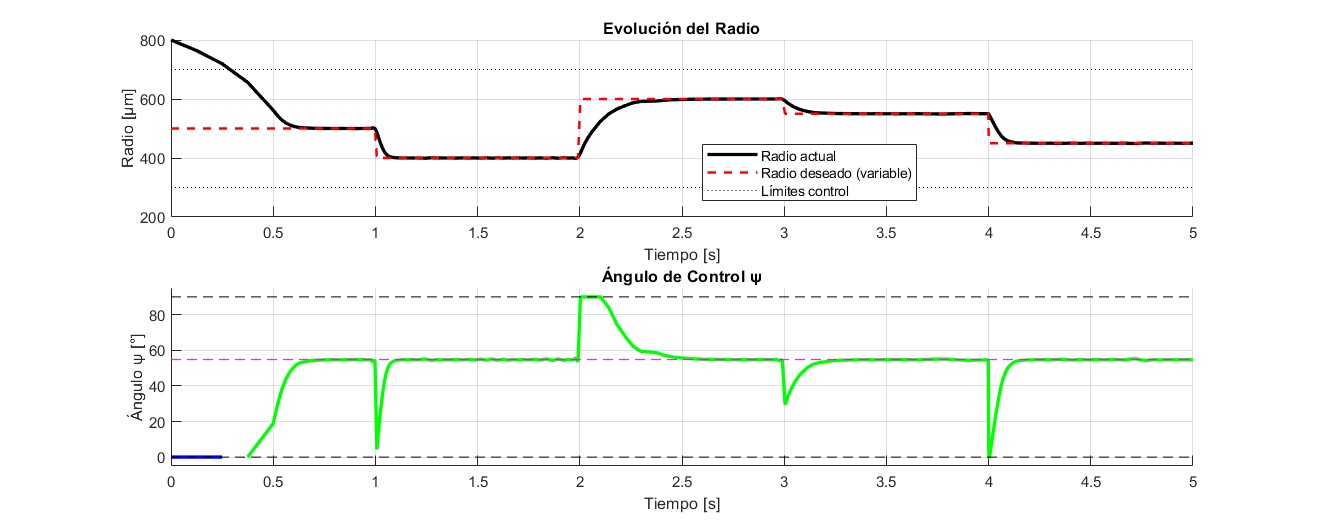}
    \caption{Simulation of PID control applied to multiple target radii}
    \label{fig:radiosmultiplesPID}
\end{figure}

Figures \ref{fig:radiosmultiplesP} and \ref{fig:radiosmultiplesPID} show the results. 

The PID controller adapts quickly to each new target, accurately reaching all of them. In contrast, during the first change, the proportional controller fails to reach the target before the next update, demonstrating the PID’s advantage in dynamic conditions.\\

However, this improvement in precision and response speed introduces a new challenge related to the rate of change of the field angle. In particular, the PID controller produces abrupt angular variations. The simulation does not account for the actual physical time required for the microrobot to rotate, nor for mechanical constraints on such rapid reorientations.\\

To address this, a second controller was designed to smooth the angular evolution, ensuring a continuous and realistic trajectory.

\subsubsection{Cascade}

To smooth these abrupt angular changes, a PD controller was implemented (Appendix A.\ref{lst:codigoPIDmultiplePD}), which considers both the angle computed by the previous controller and the angle applied in the previous time step. This second control layer is executed immediately after the PID, serving as an additional stage to improve signal continuity.\\

The integral term was omitted, as steady-state error is not a major concern here, and its inclusion could introduce unnecessary instability.

\begin{figure}[H]
    \centering
    \includegraphics[width=\linewidth]{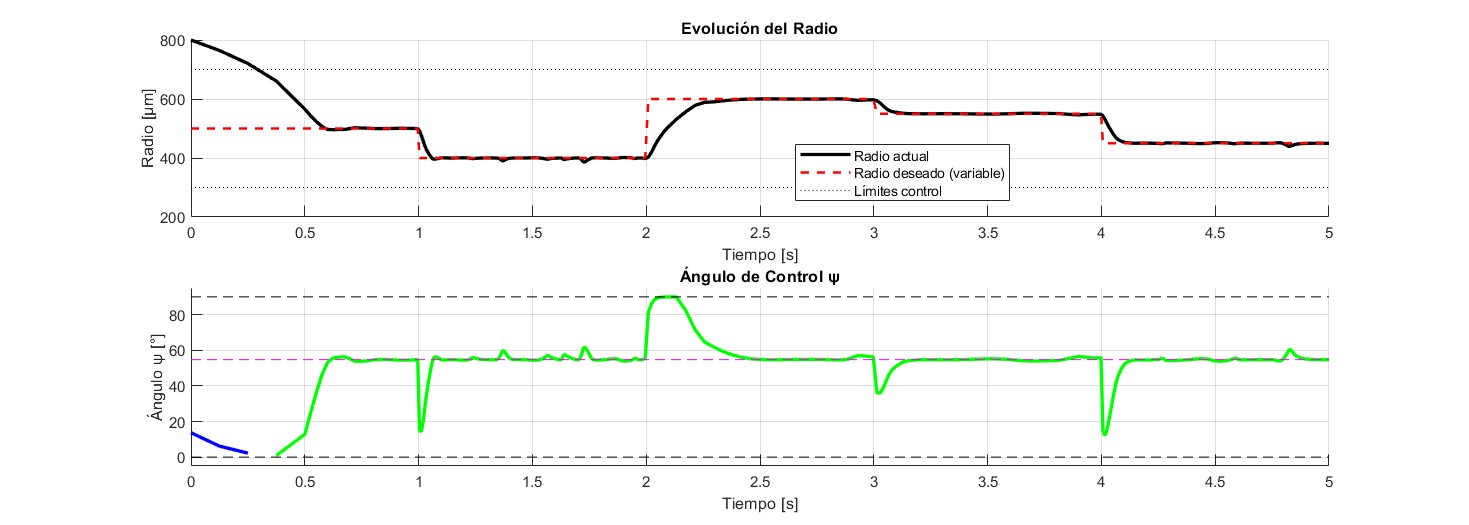}
    \caption{Simulation of the cascade controller (PID + PD) applied to multiple target radii}
    \label{fig:controladorcadena}
\end{figure}

As shown in Figure \ref{fig:controladorcadena}, while the angle still fluctuates, these variations are substantially smoother compared to using only the PID controller.\\

The pronounced peaks were attenuated, reducing abrupt changes and avoiding extreme angles ($0^{\circ}$ or $90^{\circ}$), resulting in a more realistic and feasible control scheme for future physical applications. The additional controller did not significantly increase computation time.\\

Although the radius presents slight fluctuations, they remain minimal.\\

A classic control architecture combining PID and PD was selected mainly for its straightforward implementation, intuitive physical interpretation, and robust performance in systems with non-perfectly-linear dynamics such as this one.\\

Specifically, the PID controller efficiently corrects errors by balancing response speed, precision, and stability, while the cascade PD provides smoothing of the control signal without error accumulation.\\

Although more advanced alternatives - such as nonlinear or optimal control strategies - exist, they require more complex modeling and precise parameter estimation, which would deviate from this study’s original goal: to identify a functional, simple, and robust control strategy for an underactuated system under idealized conditions.\\

\subsection{Future Work}
While the proposed controller performs well under the analyzed conditions, its applicability is limited to 2D systems and considers only the angle $\psi$ as a control parameter acting on two agents. As the number of agents increases, controller effectiveness decreases due to stronger coupling effects and underactuation.\\

Extending this controller is non-trivial, as the coupling between agents is nonlinear, and multi-agent interactions become increasingly complex, requiring a deeper analysis of the system.\\

Additionally, this model assumes that the normal motion component remains constant, which is valid only under the specific conditions of the medium - assumed here to be honey. \\

Therefore, it is proposed to study the system’s sensitivity to varying viscosities, both constant and time-varying, through simulations replicating real biological fluids. This would enable evaluating controller robustness under diverse physical conditions.\\

Further simulations should also include external disturbances, allowing the controller to be tested for noise resistance.\\

Future research will focus on extending the controller to three-dimensional systems with multiple agents, including obstacle interaction and corresponding simulations. Finally, experimental validation with real agents is planned to assess performance in practical biomedical applications.

\section{Conclusion}

This research demonstrated the feasibility of controlling the radial distance between two agents in 2D using a shared global magnetic field, thus validating the initial hypothesis. The results showed that the PID controller reduced convergence time to the target radius by approximately 40\% compared to proportional control.\\

Adding the cascade PD controller produced smoother angular trajectories, reducing abrupt variations to within $\pm5^{\circ}$ with minimal radial deviations, making it suitable for physical implementation.\\

This control strategy is particularly relevant for biomedical applications requiring micrometric precision in multi-agent control, such as targeted drug delivery and minimally invasive microsurgery.\\ 

Although limited to 2D and two agents, the proposed model establishes a solid foundation for extending to 3D environments with variable viscosity and larger agent populations. It represents a well-grounded advancement in the control of underactuated microrobotic systems.

\newpage
\section*{Glossary}
The following glossary defines the main physical and control-related concepts used throughout this report:

\begin{enumerate}
    \item \textbf{APPLIED MAGNETIC FIELD ($\mathbf{b_a}$):} External signal that orients the magnetic moments of the microrobots. It is modulated by the angles $\psi$ and $\alpha$ (\ref{campo}).

    \item \textbf{PID CONTROLLER:} Control system that combines Proportional, Integral, and Derivative actions to adjust variables such as radial distance $(r)$. 

    \item \textbf{MAGNETIC DIPOLE:} Physical model representing the microrobots, characterized by their magnetic moment $(m)$. It generates attractive or repulsive forces depending on its orientation (\ref{campomagentico}{}).

    \item \textbf{RADIAL FORCE $(f_r)$:} Component of the force between microrobots that affects their distance. It depends on the angle $\psi$ (\ref{fuerzaradial}).

    \item \textbf{STOKES' LAW:} Describes the motion of spheres in viscous fluids. Used to calculate radial (\ref{stokesradial}) and angular velocities (\ref{angularstokes}).

    \item \textbf{VISCOUS MEDIUM:} Fluid (e.g., honey, $\mu_m = 0.5\,[Pa\cdot s]$) where the microrobots operate, determining their dynamic friction ($\sigma_{\text{tras}}, \sigma_{\text{rot}}$ in Table \ref{tab:parametros}).

    \item \textbf{MAGNETIC MOMENT $(m)$:} Vector property that quantifies the magnetic strength and orientation of a microrobot ($6.545\cdot 10^{-7}\,[A·m^2]$ in this research).

    \item \textbf{MAGNETIC PERMEABILITY $(\mu)$:} Ability of a medium (e.g., honey) to support the formation of magnetic fields. In this research, $\mu \approx \mu_0$ (\ref{permeabilidad}).

    \item \textbf{UNDERACTUATION:} Limitation where the number of control parameters ($\psi$) is smaller than the system’s degrees of freedom ($r, \phi$).
\end{enumerate}

\newpage

\section*{Acknowledgements}
I would like to express my gratitude to Professor Alejandro Maass for his guidance and for opening the doors to this research, as well as to Santiago Gomez, whose initial advice was key to contacting the professor and initiating this project.\\

My gratitude also goes to Michel Rozas, whose enthusiasm for electrical engineering inspired me to pursue this path, and whose help in understanding certain concepts of this research offered a new perspective on our work.\\

I would also like to thank my father, Guillermo Fernández-Bunster, whose support in writing was invaluable in bringing clarity and coherence to each section.\\

Finally, I am grateful to the UC Engineering Directorate for Research and Innovation for promoting spaces such as IPre, where students can explore research beyond the visible curriculum and delve into more specialized topics.

\newpage
\printbibliography{} 

\newpage
\appendix
\section{Source code}

Variable names remain in Spanish to preserve consistency with the original research workflow, which was initially conducted in Spanish and later translated into English.

\subsection{MATLAB implementation of the P controller (single target)}

\begin{lstlisting}[language=Matlab, 
                 caption={MATLAB implementation of the PID controller (single target)}, 
                 label={lst:codigoP}]
function fuera_rango_mayor_P() %P
    % Basic parameters
    R = 250e-6;           % Microrobot radius [m]
    m = 6.545e-7;         % Magnetic moment [A*m^2]
    mu_m = 0.5;           % Medium viscosity [Pa*s]

    % Calculated constants
    Omega_t = 2.7271e-17 / mu_m;  % [HA^2m^2]

    error_r = 0;
    error_anterior = 0;
    psi_i = 0;

    % Control parameters
    r_deseado = 500e-6;   % Desired radius [m]
    r_min = 300e-6;       % Lower limit [m]
    r_max = 700e-6;       % Upper limit [m]
    K = 0.35;             % Controller gain
    psi_s = 54.74;        % Equilibrium angle [degrees]

    % Initial conditions
    r0 = 800e-6;          % Initial radius [m] (out of range)
    tspan = [0 1];        % Simulation time [s]

    % Solve the dynamics
    [t, r] = ode45(@(t,r) dynamics(t, r, Omega_t, r_deseado, r_min, r_max, K, psi_i, error_anterior, error_r), tspan, r0);

    % Compute psi at each step
    psi = zeros(size(r));
    tipo = zeros(size(r));  % Store control type
    for i = 1:length(r)
        [psi(i), tipo(i)] = calculate_psi(r(i), r_deseado, r_min, r_max, K, psi_i, error_anterior, error_r);
    end

    % Plots
    figure;

    % 1. Radius plot
    subplot(2,1,1);
    hold on;
    plot(t, r*1e6, 'k', 'LineWidth', 2);  % Black solid line
    plot(t, r_deseado*1e6*ones(size(t)), 'k--');
    plot(t, r_min*1e6*ones(size(t)), 'k:');
    plot(t, r_max*1e6*ones(size(t)), 'k:');
    xlabel('Time [s]');
    ylabel('Radius [mu*m]');
    title('Radius evolution');
    legend('Current radius', 'Desired radius', 'Control limits', 'Location', 'best');
    grid on;
    hold off;

    % 2. Psi angle plot
    subplot(2,1,2);
    hold on;

    % Light red semi-transparent background where psi < 0 or psi > 90
    y_min = -5;
    y_max = 95;
    mask = (psi < 0) | (psi > 90);
    in_zone = false;
    start_idx = 0;

    for i = 1:length(mask)
        if mask(i) && ~in_zone
            start_idx = i;
            in_zone = true;
        elseif ~mask(i) && in_zone
            end_idx = i - 1;
            fill([t(start_idx:end_idx); flipud(t(start_idx:end_idx))], ...
                 [y_min*ones(end_idx-start_idx+1,1); y_max*ones(end_idx-start_idx+1,1)], ...
                 [1, 0.8, 0.8], 'EdgeColor', 'none', 'FaceAlpha', 0.5);
            in_zone = false;
        end
    end
    if in_zone
        end_idx = length(t);
        fill([t(start_idx:end_idx); flipud(t(start_idx:end_idx))], ...
             [y_min*ones(end_idx-start_idx+1,1); y_max*ones(end_idx-start_idx+1,1)], ...
             [1, 0.8, 0.8], 'EdgeColor', 'none', 'FaceAlpha', 0.5);
    end

    % Line plotting
    for i = 1:3
        idx = tipo == i;
        switch i
            case 1
                plot(t(idx), psi(idx), 'Color', [1, 0.5, 0], 'LineWidth', 2); % orange
            case 2
                plot(t(idx), psi(idx), 'b', 'LineWidth', 2); % blue
            case 3
                plot(t(idx), psi(idx), 'g', 'LineWidth', 2); % green
        end
    end
    plot(t, 0*ones(size(t)), 'k--');
    plot(t, 90*ones(size(t)), 'k--');
    plot(t, psi_s*ones(size(t)), 'm--');
    xlabel('Time [s]');
    ylabel('Control angle psi [degrees]');
    title('Control Angle psi');
    ylim([y_min y_max]);
    grid on;
    hold off;
end

function drdt = dynamics(t, r, Omega_t, r_deseado, r_min, r_max, K, psi_i, error_anterior, error_r)
    [psi, ~] = calculate_psi(r, r_deseado, r_min, r_max, K, psi_i, error_anterior, error_r);
    psi_rad = deg2rad(psi);
    drdt = (Omega_t/r^4) * (1 - 3*(cos(psi_rad))^2);
end

function [psi, type] = calculate_psi(r, r_deseado, r_min, r_max, K, psi_i, error_anterior, error_r)
    kp = 0.2;
    ki = 0;
    kd = 0;
    if r < r_min
        psi = 90;
        type = 1;
    elseif r > r_max
        psi = 0;
        type = 2;
    else
        error_anterior = error_r;
        error_r = (r - r_deseado)*1e6;
        psi_p = kp * error_r;
        psi_i = psi_i + ki * error_r;
        psi_d = kd * (error_r - error_anterior);
    
        psi = 54.74 - (psi_p + psi_d + psi_i);
        type = 3;
        psi = min(max(0,psi),90);
    end
end
\end{lstlisting}

\subsection{MATLAB implementation of the PID controller (single target)}

\begin{lstlisting}[language=Matlab, 
                 caption={MATLAB implementation of the PID controller (single target)}, 
                 label={lst:codigoPID}]
function fuera_rango_mayor_PID() %PID
    % Basic parameters
    R = 250e-6;           % Microrobot radius [m]
    m = 6.545e-7;         % Magnetic moment [A*m^2]
    mu_m = 0.5;           % Medium viscosity [Pa*s]

    % Calculated constants
    Omega_t = 2.7271e-17 / mu_m;  % [HA^2m^2]

    error_r = 0;
    error_anterior = 0;
    psi_i = 0;

    % Control parameters
    r_deseado = 500e-6;   % Desired radius [m]
    r_min = 300e-6;       % Lower limit [m]
    r_max = 650e-6;       % Upper limit [m]
    K = 0.35;             % Controller gain
    psi_s = 54.74;        % Equilibrium angle [degrees]

    % Initial conditions
    r0 = 800e-6;          % Initial radius [m] (out of range)
    tspan = [0 1];        % Simulation time [s]

    % Solve dynamics
    [t, r] = ode45(@(t,r) dynamics(t, r, Omega_t, r_deseado, r_min, r_max, K, psi_i, error_anterior, error_r), tspan, r0);

    % Compute psi at each step
    psi = zeros(size(r));
    tipo = zeros(size(r));  % Store control type
    for i = 1:length(r)
        [psi(i), tipo(i)] = calculate_psi(r(i), r_deseado, r_min, r_max, K, psi_i, error_anterior, error_r);
    end

    % Plots
    figure;

    % 1. Radius plot
    subplot(2,1,1);
    hold on;
    plot(t, r*1e6, 'k', 'LineWidth', 2);  % Solid black line
    plot(t, r_deseado*1e6*ones(size(t)), 'k--');
    plot(t, r_min*1e6*ones(size(t)), 'k:');
    plot(t, r_max*1e6*ones(size(t)), 'k:');
    xlabel('Time [s]');
    ylabel('Radius [mu*m]');
    title('Radius Evolution');
    legend('Current radius', 'Desired radius', 'Control limits', 'Location', 'best');
    grid on;
    hold off;

    % 2. Psi angle plot
    subplot(2,1,2);
    hold on;

    % Light red semi-transparent background where psi < 0 or psi > 90
    y_min = -5;
    y_max = 95;
    mask = (psi < 0) | (psi > 90);
    in_zone = false;
    start_idx = 0;

    for i = 1:length(mask)
        if mask(i) && ~in_zone
            start_idx = i;
            in_zone = true;
        elseif ~mask(i) && in_zone
            end_idx = i - 1;
            fill([t(start_idx:end_idx); flipud(t(start_idx:end_idx))], ...
                 [y_min*ones(end_idx-start_idx+1,1); y_max*ones(end_idx-start_idx+1,1)], ...
                 [1, 0.8, 0.8], 'EdgeColor', 'none', 'FaceAlpha', 0.5);
            in_zone = false;
        end
    end
    if in_zone
        end_idx = length(t);
        fill([t(start_idx:end_idx); flipud(t(start_idx:end_idx))], ...
             [y_min*ones(end_idx-start_idx+1,1); y_max*ones(end_idx-start_idx+1,1)], ...
             [1, 0.8, 0.8], 'EdgeColor', 'none', 'FaceAlpha', 0.5);
    end

    % Line plotting
    for i = 1:3
        idx = tipo == i;
        switch i
            case 1
                plot(t(idx), psi(idx), 'Color', [1, 0.5, 0], 'LineWidth', 2); % orange
            case 2
                plot(t(idx), psi(idx), 'b', 'LineWidth', 2); % blue
            case 3
                plot(t(idx), psi(idx), 'g', 'LineWidth', 2); % green
        end
    end
    plot(t, 0*ones(size(t)), 'k--');
    plot(t, 90*ones(size(t)), 'k--');
    plot(t, psi_s*ones(size(t)), 'm--');
    xlabel('Time [s]');
    ylabel('Control angle psi [degrees]');
    title('Control Angle psi');
    ylim([y_min y_max]);
    grid on;
    hold off;
end

function drdt = dynamics(t, r, Omega_t, r_deseado, r_min, r_max, K, psi_i, error_anterior, error_r)
    [psi, ~] = calculate_psi(r, r_deseado, r_min, r_max, K, psi_i, error_anterior, error_r);
    psi_rad = deg2rad(psi);
    drdt = (Omega_t/r^4) * (1 - 3*(cos(psi_rad))^2);
end

function [psi, type] = calculate_psi(r, r_deseado, r_min, r_max, K, psi_i, error_anterior, error_r)
    kp = 0.2;
    ki = 0.18;
    kd = 0.2;
    if r < r_min
        psi = 90;
        type = 1;
    elseif r > r_max
        psi = 0;
        type = 2;
    else
        error_anterior = error_r;
        error_r = (r - r_deseado)*1e6;
        psi_p = kp * error_r;
        psi_i = psi_i + ki * error_r;
        psi_d = kd * (error_r - error_anterior);
    
        psi = 54.74 - (psi_p + psi_d + psi_i);
        type = 3;
        psi = min(max(0,psi),90);
    end
end
\end{lstlisting}

\subsection{MATLAB implementation of the PID controller with multiple target radii}

\begin{lstlisting}[language=Matlab, 
                 caption={MATLAB implementation of the PID controller with multiple target radii}, 
                 label={lst:codigoPIDmultiple}]
function radios_cambiantes_PID() % Multiple-radius PID controller
    % Basic parameters
    R = 250e-6;           % Microrobot radius [m]
    m = 6.545e-7;         % Magnetic moment [A*m^2]
    mu_m = 0.5;           % Medium viscosity [Pa*s]

    % Calculated constants
    Omega_t = 2.7271e-17 / mu_m;  % [HA^2*m^2]

    error_r = 0;
    error_anterior = 0;
    psi_i = 0;

    % Control parameters
    r_deseado_inicial = 500e-6;   % Initial desired radius [m]
    r_min = 300e-6;               % Lower limit [m]
    r_max = 700e-6;               % Upper limit [m]
    K = 0.35;                     % Controller gain
    psi_s = 54.74;                % Equilibrium angle [degrees]

    % Initial conditions
    r0 = 800e-6;                  % Initial radius [m] (outside range)
    tspan = [0 5];                % Simulation time [s] (extended to observe multiple changes)
    
    % Define desired radius changes every second
    r_deseado_valores = [500e-6, 400e-6, 600e-6, 550e-6, 450e-6]; % Desired radius sequence
    
    % Solve system dynamics with varying desired radius
    options = odeset('OutputFcn', @odeplot);
    [t, r] = ode45(@(t,r) dynamics(t, r, Omega_t, get_current_r_deseado(t, r_deseado_inicial, r_deseado_valores),...
                                  r_min, r_max, K, psi_i, error_anterior, error_r), tspan, r0);

    % Compute psi for each step
    psi = zeros(size(r));
    tipo = zeros(size(r));          % Store control type
    r_deseado_actual = zeros(size(r)); % Store current desired radius
    for i = 1:length(r)
        current_time = t(i);
        current_r_deseado = get_current_r_deseado(current_time, r_deseado_inicial, r_deseado_valores);
        r_deseado_actual(i) = current_r_deseado;
        [psi(i), tipo(i)] = calculate_psi(r(i), current_r_deseado, r_min, r_max, K, psi_i, error_anterior, error_r);
    end

    % ----- Plot results -----
    figure;

    % 1. Radius evolution plot
    subplot(2,1,1);
    hold on;
    plot(t, r*1e6, 'k', 'LineWidth', 2);                   % Solid black line
    plot(t, r_deseado_actual*1e6, 'r--', 'LineWidth', 1.5); % Red dashed line for variable desired radius
    plot(t, r_min*1e6*ones(size(t)), 'k:');
    plot(t, r_max*1e6*ones(size(t)), 'k:');
    xlabel('Time [s]');
    ylabel('Radius [mu*m]');
    title('Radius Evolution');
    legend('Current radius', 'Desired radius (variable)', 'Control limits', 'Location', 'best');
    grid on;
    hold off;

    % 2. psi (control angle) plot
    subplot(2,1,2);
    hold on;

    % Light red shaded background where psi < 0 or psi > 90
    y_min = -5;
    y_max = 95;
    mask = (psi < 0) | (psi > 90);
    in_zone = false;
    start_idx = 0;

    for i = 1:length(mask)
        if mask(i) && ~in_zone
            start_idx = i;
            in_zone = true;
        elseif ~mask(i) && in_zone
            end_idx = i - 1;
            fill([t(start_idx:end_idx); flipud(t(start_idx:end_idx))], ...
                 [y_min*ones(end_idx-start_idx+1,1); y_max*ones(end_idx-start_idx+1,1)], ...
                 [1, 0.8, 0.8], 'EdgeColor', 'none', 'FaceAlpha', 0.5);
            in_zone = false;
        end
    end
    if in_zone
        end_idx = length(t);
        fill([t(start_idx:end_idx); flipud(t(start_idx:end_idx))], ...
             [y_min*ones(end_idx-start_idx+1,1); y_max*ones(end_idx-start_idx+1,1)], ...
             [1, 0.8, 0.8], 'EdgeColor', 'none', 'FaceAlpha', 0.5);
    end

    % Line plotting
    for i = 1:3
        idx = tipo == i;
        switch i
            case 1
                plot(t(idx), psi(idx), 'Color', [1, 0.5, 0], 'LineWidth', 2); % orange
            case 2
                plot(t(idx), psi(idx), 'b', 'LineWidth', 2); % blue
            case 3
                plot(t(idx), psi(idx), 'g', 'LineWidth', 2); % green
        end
    end
    plot(t, 0*ones(size(t)), 'k--');
    plot(t, 90*ones(size(t)), 'k--');
    plot(t, psi_s*ones(size(t)), 'm--');
    xlabel('Time [s]');
    ylabel('Control angle psi [degrees]');
    title('Control Angle psi');
    ylim([y_min y_max]);
    grid on;
    hold off;
end

function drdt = dynamics(t, r, Omega_t, r_deseado, r_min, r_max, K, psi_i, error_anterior, error_r)
    [psi, ~] = calculate_psi(r, r_deseado, r_min, r_max, K, psi_i, error_anterior, error_r);
    psi_rad = deg2rad(psi);
    drdt = (Omega_t/r^4) * (1 - 3*(cos(psi_rad))^2);
end

function [psi, type] = calculate_psi(r, r_deseado, r_min, r_max, K, psi_i, error_anterior, error_r)
    kp = 0.2;
    ki = 0.18;
    kd = 0.2;
    if r < r_min
        psi = 90;
        type = 1;
    elseif r > r_max
        psi = 0;
        type = 2;
    else
        error_anterior = error_r;
        error_r = (r - r_deseado)*1e6;
        psi_p = kp * error_r;
        psi_i = psi_i + ki * error_r;
        psi_d = kd * (error_r - error_anterior);
    
        psi = 54.74 - (psi_p + psi_d + psi_i);
        type = 3;
        psi = min(max(0,psi),90);
    end
end

% Auxiliary function to obtain the current desired radius based on time
function r_deseado = get_current_r_deseado(t, r_deseado_inicial, r_deseado_valores)
    % Change the desired radius every second
    segundo_actual = floor(t) + 1; % +1 because MATLAB indexes from 1
    if segundo_actual > length(r_deseado_valores)
        segundo_actual = length(r_deseado_valores);
    end
    r_deseado = r_deseado_valores(segundo_actual);
end
\end{lstlisting}

\subsection{MATLAB implementation of the cascade PID+PD controller with multiple target radii}
\begin{lstlisting}[language=Matlab, 
                 caption={MATLAB implementation of the cascade PID+PD controller with multiple target radii}, 
                 label={lst:codigoPIDmultiplePD}]
function radios_cambiantes_PIDCC() % Multiple-radius Cascade PID controller
    % Basic parameters
    R = 250e-6;           % Microrobot radius [m]
    m = 6.545e-7;         % Magnetic moment [A*m^2]
    mu_m = 0.5;           % Medium viscosity [Pa*s]

    % Calculated constants
    Omega_t = 2.7271e-17 / mu_m;  % [HA^2*m^2]

    error_r = 0;
    error_anterior = 0;
    psi_i = 0;

    % Control parameters
    r_deseado_inicial = 500e-6;   % Initial desired radius [m]
    r_min = 300e-6;               % Lower limit [m]
    r_max = 700e-6;               % Upper limit [m]
    K = 0.35;                     % Controller gain
    psi_s = 54.74;                % Equilibrium angle [degrees]

    % Initial conditions
    r0 = 800e-6;                  % Initial radius [m] (outside range)
    tspan = [0 5];                % Extended simulation time to observe transitions
    
    % Define desired radius changes every second
    r_deseado_valores = [500e-6, 400e-6, 600e-6, 550e-6, 450e-6]; % Desired radius sequence
    
    % Solve dynamics with changing desired radius
    options = odeset('OutputFcn', @odeplot);
    [t, r] = ode45(@(t,r) dynamics(t, r, Omega_t, get_current_r_deseado(t, r_deseado_inicial, r_deseado_valores),...
                                  r_min, r_max, K, psi_i, error_anterior, error_r), tspan, r0);

    % Compute psi for each time step
    psi = zeros(size(r));
    tipo = zeros(size(r));            % Store control type
    r_deseado_actual = zeros(size(r)); % Store current desired radius
    for i = 1:length(r)
        current_time = t(i);
        current_r_deseado = get_current_r_deseado(current_time, r_deseado_inicial, r_deseado_valores);
        r_deseado_actual(i) = current_r_deseado;
        [psi(i), tipo(i)] = calculate_psi(r(i), current_r_deseado, r_min, r_max, K, psi_i, error_anterior, error_r);
    end

    % ----- Plot results -----
    figure;

    % 1. Radius evolution plot
    subplot(2,1,1);
    hold on;
    plot(t, r*1e6, 'k', 'LineWidth', 2);                    % Solid black line
    plot(t, r_deseado_actual*1e6, 'r--', 'LineWidth', 1.5); % Red dashed line for variable desired radius
    plot(t, r_min*1e6*ones(size(t)), 'k:');
    plot(t, r_max*1e6*ones(size(t)), 'k:');
    xlabel('Time [s]');
    ylabel('Radius [mu*m]');
    title('Radius Evolution');
    legend('Current radius', 'Desired radius (variable)', 'Control limits', 'Location', 'best');
    grid on;
    hold off;

    % 2. Control angle psi plot
    subplot(2,1,2);
    hold on;

    % Light red shaded background where psi < 0 or psi > 90
    y_min = -5;
    y_max = 95;
    mask = (psi < 0) | (psi > 90);
    in_zone = false;
    start_idx = 0;

    for i = 1:length(mask)
        if mask(i) && ~in_zone
            start_idx = i;
            in_zone = true;
        elseif ~mask(i) && in_zone
            end_idx = i - 1;
            fill([t(start_idx:end_idx); flipud(t(start_idx:end_idx))], ...
                 [y_min*ones(end_idx-start_idx+1,1); y_max*ones(end_idx-start_idx+1,1)], ...
                 [1, 0.8, 0.8], 'EdgeColor', 'none', 'FaceAlpha', 0.5);
            in_zone = false;
        end
    end
    if in_zone
        end_idx = length(t);
        fill([t(start_idx:end_idx); flipud(t(start_idx:end_idx))], ...
             [y_min*ones(end_idx-start_idx+1,1); y_max*ones(end_idx-start_idx+1,1)], ...
             [1, 0.8, 0.8], 'EdgeColor', 'none', 'FaceAlpha', 0.5);
    end

    % Line plotting by control type
    for i = 1:3
        idx = tipo == i;
        switch i
            case 1
                plot(t(idx), psi(idx), 'Color', [1, 0.5, 0], 'LineWidth', 2); % orange
            case 2
                plot(t(idx), psi(idx), 'b', 'LineWidth', 2); % blue
            case 3
                plot(t(idx), psi(idx), 'g', 'LineWidth', 2); % green
        end
    end
    plot(t, 0*ones(size(t)), 'k--');
    plot(t, 90*ones(size(t)), 'k--');
    plot(t, psi_s*ones(size(t)), 'm--');
    xlabel('Time [s]');
    ylabel('Control angle psi [degrees]');
    title('Control Angle psi');
    ylim([y_min y_max]);
    grid on;
    hold off;
end

function drdt = dynamics(t, r, Omega_t, r_deseado, r_min, r_max, K, psi_i, error_anterior, error_r)
    [psi, ~] = calculate_psi(r, r_deseado, r_min, r_max, K, psi_i, error_anterior, error_r);
    psi_rad = deg2rad(psi);
    drdt = (Omega_t/r^4) * (1 - 3*(cos(psi_rad))^2);
end

function [psi, type] = calculate_psi(r, r_deseado, r_min, r_max, K, psi_i, error_anterior, error_r)
    persistent psi_aplicado psi_i_c e_anterior psi_anterior e
    if isempty(psi_aplicado)
        psi_aplicado = 54.74;
        psi_i_c = 0;
        e_anterior = 0;
        psi_anterior = 54.74;
        e = 0;
    end

    % --- PID to compute target psi ---
    kp = 0.2;
    ki = 0.18;
    kd = 0.2;

    if r < r_min
        psi_objetivo = 90;
        type = 1;
    elseif r > r_max
        psi_objetivo = 0;
        type = 2;
    else
        error_anterior = error_r;
        error_r = (r - r_deseado)*1e6;
        psi_p = kp * error_r;
        psi_i = psi_i + ki * error_r;
        psi_d = kd * (error_r - error_anterior);

        psi_objetivo = 54.74 - (psi_p + psi_d + psi_i);
        type = 3;
    end
    psi_objetivo = min(max(0, psi_objetivo), 90);

    % --- Cascade PID to smooth psi response ---
    kpc = 0.7;
    kic = 0;
    kdc = 0.05;
    
    e_anterior = e;
    e = psi_aplicado - psi_objetivo;
    psi_p_c = kpc * e;
    psi_i_c = psi_i_c + kic * e;
    psi_d_c = kdc * (e - e_anterior);

    psi_aplicado = psi_aplicado - (psi_p_c + psi_d_c + psi_i_c);
    psi_aplicado = min(max(0, psi_aplicado), 90);

    % Update internal state
    e_anterior = e;
    psi_anterior = psi_aplicado;

    psi = psi_aplicado;
end

% Auxiliary function to obtain the current desired radius based on time
function r_deseado = get_current_r_deseado(t, r_deseado_inicial, r_deseado_valores)
    % Change the desired radius every second
    segundo_actual = floor(t) + 1; % +1 because MATLAB indexes from 1
    if segundo_actual > length(r_deseado_valores)
        segundo_actual = length(r_deseado_valores);
    end
    r_deseado = r_deseado_valores(segundo_actual);
end
\end{lstlisting}
\end{document}